# Is China Entering WTO or *shijie maoyi zuzhi*?
## —a Corpus Study of English Acronyms in Chinese Newspapers[1]

## Hai Hu
*Indiana University Bloomington*

This is one of the first studies that quantitatively examine the usage of English acronyms (e.g. WTO) in Chinese texts. Using newspaper corpora, I try to answer 1) for all instances of a concept that has an English acronym (e.g. *World Trade Organization*), what percentage is expressed in the English acronym (WTO), and what percentage in its Chinese translation (世界贸易组织 *shijie maoyi zuzhi*), and 2) what factors are at play in language users' choice between the English and Chinese forms? Results show that different concepts have different percentage for English acronyms (*PercentOfEn*), ranging from 2% to 98%. Linear models show that *PercentOfEn* for individual concepts can be predicted by language economy (how long the Chinese translation is), concept frequency, and whether the first appearance of the concept in Chinese newspapers is the English acronym or its Chinese translation (all $p < .05$).

## 0. Introduction

As one of the oldest languages in the world, Mandarin Chinese is reluctant to borrow words from other languages. It is ranked as the lowest borrower on a list 41 languages in the *Loanword Typology Project* (LWT, Haspelmath & Tadmor 2009: 56-57). As comprehensive as it strives to be, the LWT project looked only at a total of 1460 *core* meanings across languages. The focus of LWT is not to examine how the language deals with very recent words and concepts originated from other languages, but this has become a major issue in language contact in recent years. In China, there has been a heated debate since 2000 both in the academia and mass media on whether to allow common English acronyms, e.g. WTO, NBA, GDP, etc. to appear untranslated in Chinese text (for a review, see Yu & Zhu 2003).

Despite the lively debate among scholars and language policy makers, one fundamental question has not been fully explored. That is, what exactly is the current linguistic situation? Without a clear answer to the degree of the English acronyms "invasion", any conclusion would seem hasty. Thus the first research question of the current paper is:

---

[1] I am very grateful to Michael Frisby at Indiana Statistical Consulting Center for his help with the statistical analysis. I also thank Prof. Clancy Clements and Prof. Markus Dickinson for helpful discussions. The author is funded by China Scholarship Council.

1) What is the percentage of English acronym vs. its Chinese translation for a given concept in Chinese newspapers?

For instance, what percentage of the concept *World Trade Organization* is expressed in English acronym (WTO), and what percentage in Chinese (世界贸易组织 *shijie maoyi zuzhi*, literally meaning *World Trade Organization*). The paper tries to answer this question by examining: i) the usage of 10 concepts such as WTO, GDP, CEO, etc. in three Chinese newspapers (*People's Daily*, *China Business News*, and *Southern Metropolis Daily*) in a time span of ten years, and ii) another corpus of Xinhua News where the usage of a larger set of 31 concepts are explored.

Next, linear models (following Zenner et al. 2014) are used to answer the second research question of the paper:

2) What factors influence the percentage of English acronyms in Chinese newspapers? That is, what factors can help us predict how much of a concept will appear in the English form?

To preview the results, more formal newspapers tend to use fewer English acronyms; proper nouns are more likely to be used as an English acronym; speech economy (i.e. the length of the Chinese translation) is the most important predictor for acronym percentage.

Following this section of introduction is a brief literature review. Then section 2 introduces the methods, English acronyms and corpora used in the paper. Section 3 reports the results for the two research questions and an interview with a newspaper editor, and discusses the implications of the results. Section 4 concludes the paper.

## 1. Literature Review

Recently, researchers have studied the influence of English on Chinese lexical affixation (Wu 2001), syntax of written Chinese (Shi & Chu 1999), as well as how the English words are translated to fit the phonotactics of Chinese (Hall-Lew 2002: 26-28). Here we briefly review the studies relevant to lexical borrowing.

In a sociolinguistic study by Hall-Lew (2002), she reports that Chinese speakers use English words/acronyms in their speech because "these forms were often 'easier' to say than the Chinese loans [i.e the Chinese translation]". Her informants also suggest that the English words are mostly used in "casual conversation" (2002: 28).

Yu & Zhu (2002) provides three reasons for the popular use of English acronyms in Chinese conversations and written texts. The macro-political-economic superiority of the English language is one of the important reasons. But more interestingly, they argue that Chinese speakers find the English acronym WTO "simpler and more concise", as well as "more novel" compared to the Chinese translation "世界贸易组织".

These studies suggest that English acronyms are preferred for reasons of *language economy* (WTO is simpler and more concise) and *formality* (English words and acronyms appear often in casual situations). By language (or speech) economy I mean the principle that speakers often prefer shorter, simpler words while other conditions being roughly equal. These two reasons will be reflected in the choice of predictors in linear models in Section 2.3, which also take into account the predictors proposed in Zenner et al.'s paper (2014) on loanwords and borrowability in Dutch.

In their study, Zenner et al. (2014) explores the correlation between the *coreness* of a concept and its borrowability in Dutch, using a large corpora of 11 newspapers. Specifically, they used a statistical model to examine whether the English form or the Dutch form is used more frequently. That is, for a concept like BACKPACKER that has both English and Dutch forms, which appears more frequently: the English form *backpacker* or its Dutch counterpart *rugzakker*? Their results show that the higher the frequency of the concept, the less likely that the English form will be used; also, the older the concept, the lower the percentage of the English word. According to them, the frequency and age of a concept is the measurement of *coreness* of a concept: higher frequency and longer age indicates more *coreness*. Another predictor, speech economy— whether the foreign word has the smallest number of alphabets compared to all other possible English and Dutch forms—only plays a significant role under certain conditions.

One goal of the current study is to examine how well the above-mentioned predictors can predict the percentage of English acronym usage, which is analyzed in detail in Section 3.2.

## 2. Methodology

In this section, I describe the English acronyms and newspaper corpora used in the study and the linear models that predict the percentage of English acronyms.

### 2.1 Acronyms Examined in the Study

The first part of study examines the usage of 10 economy/finance related acronyms in three newspaper corpora (see Table 1). These acronyms are chosen from the Appendix of *Contemporary Chinese Dictionary 6th Edition* (2012)[2].

One reason for choosing words from the economy sector is that there are more English acronyms in this sector whose Chinese translation has varying number of characters, which makes it easier to test whether language economy (as the number of characters in Chinese) plays a role in choosing between the English and the Chinese forms.

Another reason is that those acronyms listed in the Appendix of *Contemporary Chinese Dictionary 6th Edition* have been the center of the 'should we allow English acronyms' debate in the media. Some scholars even argue that it is illegal for a Chinese dictionary to include Western alphabets, even if only in the Appendix of the dictionary.[3] Yet, to my knowledge no one has examined exactly how often the acronyms in the dictionary appear. The list of 31 acronyms used for the second dataset, i.e. corpus of Xinhua News, is listed in the appendix of the paper.

---

[2] 《现代汉语词典》第六版

[3] See e.g. http://fangtan.people.com.cn/GB/147553/348511/index.html

*Table 1* Ten acronyms used for the first set of three newspaper corpora

| English acronym | English (full) | Chinese translation | Number of Characters |
|---|---|---|---|
| WTO | World Trade Organization | 世贸组织 / 世界贸易组织 | 4 / 6 |
| OPEC | Organization of the Petroleum Exporting Countries | 欧佩克 / 石油输出国组织 | 3 / 7 |
| CBD | central business district | 中央商务区 | 5 |
| CEO | chief executive officer | 首席执行官 | 5 |
| GDP | gross domestic product | 国内生产总值 | 6 |
| IPO | initial public offering | 首次公开募股 | 6 |
| MBA | Master of Business Administration | 工商管理硕士 | 6 |
| IMF | International Monetary Fund | 国际货币基金组织 | 8 |
| CPI | consumer price index | 居民消费价格指数 | 8 |
| EMBA | Executive Master of Business Administration | 高级管理人员工商管理硕士 | 12 |

## 2.2 Corpora in the Current Study
### 2.2.1 Dataset 1: Corpora of Three Newspapers

There are several considerations when choosing the newspapers. First, I intend to examine possible factors mentioned in previous literature that may influence the choice between English and Chinese versions, such as language formality (Hall-Lew 2002). Also, I hypothesized that whether or not the newspaper is a financial newspaper can also influence the usage of finance/business related acronyms. In addition, the data should be attainable, i.e. the newspaper corpora must be available. In the end, three newspapers are chosen in the current study, each with a collection spanning 10 years (2005 to 2014).

*Table 2* Four newspaper corpora used in the first part of the study—dataset 1[4]

| Newspaper | *People's Daily (PD)* | *China Business News (CBN)* | *Southern Metropolis Daily (SMD)* |
|---|---|---|---|
| **Genre** | Political | Business | General |
| **Formality** | Very formal | Formal | Less formal |
| **Publisher** | Central Committee of the Communist Party of China | China Business Network | Southern Media Group |
| **Targeted Readers** | Government officials/civil servants/the general public | Businessperson | The general public |

*People's Daily* (**PD**) is the official newspaper of the Chinese government. It is the ideal source to study formal, political language in China. It has a well-documented corpus covering news reports as early as the 1940s. *China Business Daily* (**CBN**) is a business newspaper in China, which represents the professional business/finance newspaper in the study. *Southern Metropolis Daily* (**SMD**) is a nation-wide newspaper published in the southern metropolis Guangzhou, which enjoys more liberal atmosphere and is less formal than the other two newspapers in the study.

---

[4] *PD:* 人民日报; *CBN:* 第一财经日报; *SMD:* 南方都市报.

### 2.2.2 Predictions Regarding Different Newspapers

First, I predict that the governmental newspaper, *People's Daily*, will have a lower percentage of English. The reason is that the government has discouraged the use of English acronyms. Specifically, in 2010 the State Administration of Press, Publication, Radio, Film and Television in mainland China issued a notice which forbids any English acronyms (i.e. WTO, IMF and NBA) in radio programs, TV broadcasting and possibly newspapers[5]. Therefore, it is my hope to examine the effect of such an administrative order on both *People's Daily,* which is likely to be affected more by the order and some other newspapers such as *Southern Metropolis Daily* that may be only loosely following the order. It can reveal to what degree language policy influences actual language use.

Also, it is predicted that financial newspapers are more likely to use English acronyms than general newspapers since the readers of financial papers are supposedly more familiar with the acronyms in their field.

### 2.2.3 Dataset 2: Corpus of Xinhua News

Since the above corpora are not easy to access and hence no large scale data collection can be achieved, I use another free online corpus to obtain data and test more concepts. The *Xinhua News Agency* corpus consists of news reports from Oct. 2001 to Sept. 2004 from Xinhua News Agency and is freely available from Center of Chinese Linguistics at Peking University[6]. This corpus will be used to check the conclusions from the first part of the study (see section 3.2 for detail).

*Table 3* Corpus used in the second part of the study—dataset 2

| Name of news | Publisher | Coverage | Formality |
|---|---|---|---|
| Xinhua News | Xinhua News agency | Oct. 2001 to Sept. 2004 | Formal |

### 2.3 Linear Models for Two Datasets

The current study draws upon the mixed-effects linear model used in Zenner et al (2014). This subsection gives a detailed description of the dependent variable and the predictors.

---

[5] I have not been able to find the notice *per se*, perhaps because it is a notice issued only to TV and radio stations (some say it also affects newspapers) which is not open to public. Nevertheless, multiple sources have confirmed that the notice did exist. For a summary of the related debates on the "purity" of Chinese language and the notice from the government authority, see e.g.: http://view.news.qq.com/zt/2010/CCTV_NBA/index.htm

[6] http://ccl.pku.edu.cn:8080/ccl_corpus/

**Dependent variable**: Percentage of English acronyms. This is operationalized as the log ratio between English acronym's frequency and the Chinese translation frequency.[7]

There are three **predictors** in the model for the corpora of three newspapers:

**Freq**: overall frequency of the concept in all newspapers. To use the same example, the frequency of concept CEO would be the sum of all CEO and 首席执行官 in all newspapers. This is the same as communicative entrenchment in Zenner et al. (2014).

**NumOfChar**: the number of characters in the Chinese translation. For example, the Chinese for CEO 首席执行官 has 5 Chinese characters in it. This is roughly equivalent to speech economy in Zenner et al (2014). But I consider it as a continuous variable in this study, i.e. it can be any number of characters ranging from 2 to 12 according to the data.

**FirstAppear**: this is a nominal variable which indicates whether the English and Chinese forms of the same concept enters the vocabulary at the same time. It has three values. 0 means they both enter the vocabulary at around the same time. 1 means the English form is earlier and 2 means the Chinese form is earlier.[8] The time of entering is determined by looking at the *People's Daily* corpora. As mentioned before, the *PD* corpora is the best documented in China and its content covers all issues of *PD* since 1946. Thus it is possible to track the first time a word appears in the corpus. This may not be the best solution, as it should be ideal to search the word in all newspaper corpora, but before such data are available, the *PD* corpus is the best approximate out there. Let's again take CEO as an example. The first news report containing CEO appeared on 26th April 2000, whereas its Chinese translation 首席执行官 appeared on 9th Nov. 1980. Thus the value of *FirstAppear* for CEO is 2.

Another crucial predictor in Zenner et al (2014) is concept age, i.e. the number of years from the first use of the concept to the time of newspaper publication. The idea is that the longer one concept has been in the lexicon, the more entrenched it should be, and hence less usage of the foreign form. It is only included in the linear regression model for dataset 2 because for dataset 1 most of the concepts are relatively young (< 20 years).

## 3. Results and Discussion
### 3.1 The Linguistic Landscape

Each panel in Figure 1 shows the percentage of English forms for one acronym. For example, the first panel in the second row shows that from year 2005 to year 2014, *People's Daily* (green line and triangles) has been somewhat consistent in its use of GDP vs. 国民生产总值. That is, around 55% to 75% percent of the time, *PD* uses GDP rather than 国民生产总值. After 2010, there seems to be a drop in the percentage of the use of the acronym GDP, but the trend is not very consistent, since it rises slightly in 2013 and 2014.

---

[7] For cases where all forms are in Chinese or English, we adopt a smoothing technique to make it 1% vs 99% so that it is possible to calculate the ratio. For example, EMBA is often only found in its English form. Instead of saying that it is 100% English, I rendered it 99% so that it is possible to calculate the ratio. For cases where the concept doesn't appear, the ratio is assigned to 1, i.e. the English and Chinese forms have equal frequency.

[8] The minimal year difference is 5 years. Thus if the English acronym of a concept appears in 1990 and the Chinese 1992, it will be considered at the same time.

### 3.1.1 Fewer English Acronyms in *PD*

First, the data clearly shows that the official newspaper of the government—*People's Daily* uses the most Chinese among the three. Apart from the concept EMBA, the highest percentage for English is only 80%. And four concepts—WTO, OPEC, IMF and CEO—even have a higher percentage for Chinese, especially for OPEC where a mere 2% are used in English, as shown in Figure 1. This shows that *PD* may indeed be the most conservative newspaper, insisting on using more Chinese forms. However, the financial newspaper does not use more English forms, which suggests that using more English is not considered a more professional.

*Figure 1* Percentage of English for 10 concepts in dataset 1

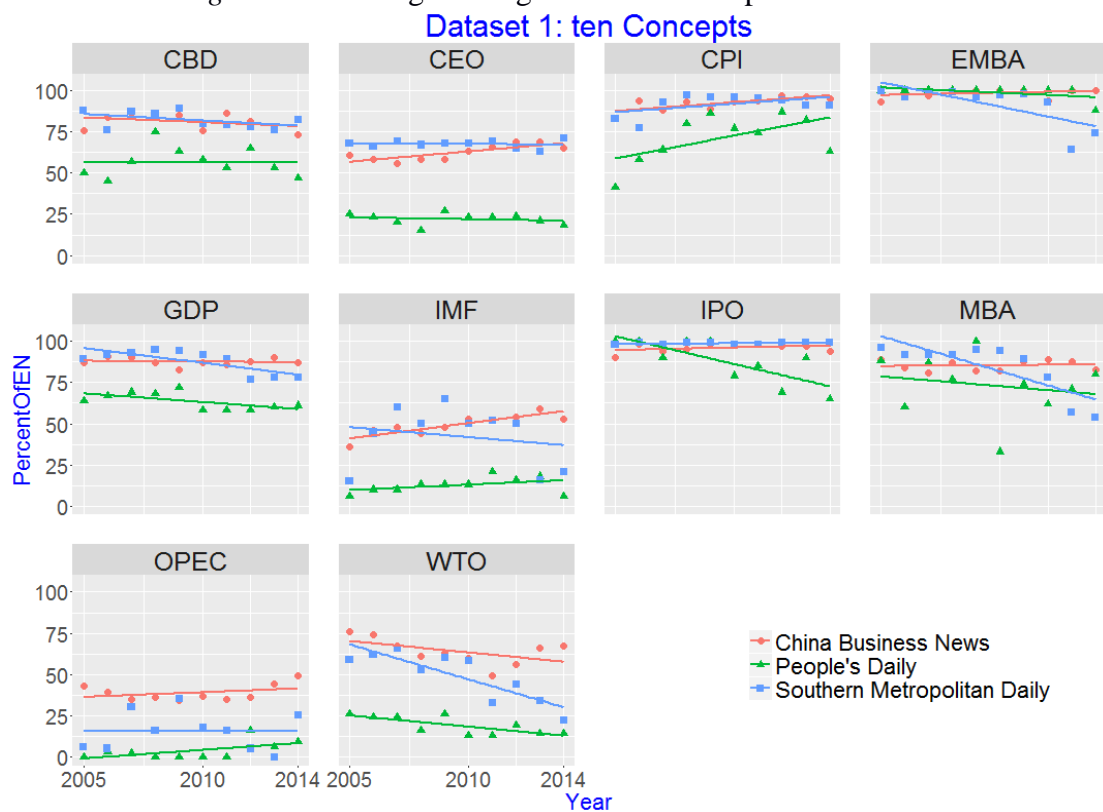

### 3.1.2 Language Policy

The initial hypothesis is that after year 2010, the percentage of English will drop because of the language policy that banned the use of English acronyms. Interestingly, *PD*, the government's official newspaper does not show a clear decline, since although for IPO and GDP the percentage of English is in decline, the percentage for CPI is rising. In contrast, the result of *Southern Metropolis Daily* (blue line and square) shows some hint of a decreasing trend, with the percentage of English dropping for EMBA, GDP, MBA and WTO. Of course, the size of the current dataset and the timespan are not enough to be conclusive, but it still implies that the effect of language policy is limited.

### 3.1.3 Proper Nouns—WTO, OPEC and IMF

An interesting thing to note is that for most concepts, English is "winning" by a large margin, except **WTO, OPEC** and **IMF**, as illustrated in Figure 1. All of them are proper nouns; specifically, they are names of an **organization**. In order to test if proper nouns have a consistent lower *PercentOfEn,* I add a new predictor to the linear regression model which considers 31 concepts. The results are reported in Section 3.2.2.

One possibility mentioned in an interview with an editor from *People's Daily* was that the *PD* readers simply may not know what organizations such as OPEC and IMF are. Therefore, the Chinese translation (国际货币基金组织, literally "international monetary fund organization") is needed to help readers understand the term. This is plausible, but it doesn't explain why non-proper-nouns such as MBA and CBD—which may still be unfamiliar to Chinese readers—are used mostly in English. Further studies with a larger corpus are needed to provide a more definitive answer.

### 3.2 Finding the Predictors

### 3.2.1 Mixed-effects Linear Model for Dataset 1—Three Newspapers

Now I report the results for the mixed-effects linear model introduced in section 2.3. To recap, we are using three predictors *NumOfChar*, *Freq*, and *FirstAppear* to predict the percentage of English acronyms *PercentOfEn* for each concept in the mixed-effects model for dataset 1. The results are shown below.

*Table 4* Results of the linear model for dataset 1

**Type III Tests of Fixed Effects[a]**

| Source | Numerator df | Denominator df | F | Sig. |
|--------|-------------|----------------|------|------|
| Intercept | 1 | 4.061 | 6.826 | .058 |
| NumOfChar | 1 | 4.512 | 14.106 | .016 |
| Freq | 1 | 4.910 | 25.825 | .004 |
| FirstAppear | 2 | 2.764 | 21.823 | .020 |

a. Dependent Variable: LogOdds1.

In this model, the newspapers are considered as subjects, and the concepts (i.e. CEO, OPEC, etc.) are taken as random effect. The results show that frequency, the number of character and whether Chinese or English enters the lexicon first all reached the level of significance (p< .05).

Further analysis shows that as the number of characters in Chinese translations increases, *PercentOfEn* also goes up, indicating that the more economic forms are preferred. In addition, the concept with higher frequencies have a higher *PercentOfEn* (which is different from the Zenner et al. 2014, see more discussion in Section 3.3). Furthermore, if the first appearance of the concept is in English, *PercentOfEn* is likely to be high. For reasons of space, I will not go into details of the analysis, but instead give a more thorough analysis for the linear regression model in the next section, which uses data from 31 concepts, rather than just 10 for dataset 1.

### 3.2.2 Linear Regression for Dataset 2—Corpus of Xinhua News

The linear regression model uses data of 31 concepts in the Xinhua News corpus to further examine the effect of the predictors. The corpus is made available by the Center for Chinese Linguistics at Peking University. Two more predictors are added. First, because of the discovery in Section 3.1 that *proper nouns* seem to have lower *PercentOfEn*, I add one predictor *ProperN* to the linear regression (which is a binary predictor, i.e. either the concept is a proper noun or it is not). Additionally, following Zenner et al. (2014), I incorporated concept age into the regression. According to them, the older the concept (the more *core* and more *entrenched* it is), the fewer English forms will be used. In this study, *ConceptAge* is also realized as a binary predictor, i.e. whether the first appearance of the concept (both English and Chinese forms count) in the *People's Daily* corpus is earlier than 1995 or not. If the concept enters *PD* before 1995, it will be considered an "old" concept. For instance, 石油输出国组织 (OPEC) is first seen in the corpus in 1960, thus making it an "old" concept, whereas IPO is "young" because it first showed up in 2003.

Thus we have the following linear regression model:

PercentOfEn ~ *NumOfChar* + *Freq* + *FirstAppear* + *ProperN* + *ConceptAge*

The goal is to see which of the five predictors can predict the percentage of English acronym of a concept. The results in Table 5 show that *NumOfChar, FirstAppear* and *Freq* are good predictors. However, *ProperN* and *ConceptAge* are not significant predictors according to the model. It also tells us that language economy, demonstrated in *NumOfChar* plays a major role in speakers' choices (p < .001), which will be further explained later. We also see that *Freq* is now in accordance with Zenner et al.'s prediction, i.e. an inverse proportional relationship with the percentage of English acronyms. Next, I present a detailed analysis for each predictor.

*Table 5* Results of linear regression model for dataset 2

| Model Adjusted R-square | Model sig. | *NumOfChar* | *FirstAppear* | *Freq* | *ProperN* | *ConceptAge* |
|---|---|---|---|---|---|---|
| .550 | .000 *** | .000 *** | .015 * | .046* | .062 | .744 |

### Predictor 1: NumOfChar

There is a clear trend in Figure 2 that the more characters the Chinese translation has (x-axis), the higher the percentage of English (y-axis). This confirms the hypothesis that speech economy is an important factor that can influence the choice between the English and Chinese lexical items. Thus we see in Figure 1 that for TOEFL whose Chinese translation has only two characters "托福", only about 15% of the time it is expressed as TOEFL, and for 85% of the time it is in Chinese. In contrast, EMBA with 12 characters in its translation has almost 100% *PercentOfEn*. This is also very intuitive when thinking about how much easier it is to say "我在上 EMBA 课程" rather than "我在上高级管理人员工商管理硕士课程." This finding is in accordance with the result of Zenner et al (2014) which states that speech economy plays a significant role in loanword usage in Dutch as well as interview with Chinese native speakers in Hall-Lew (2002). Nevertheless, Zenner et al (2014) suggests that speech economy is only significant under certain conditions, whereas here it is the single most important factor. The different role of speech economy may be attributed to the nature of English acronyms used in Chinese (WTO) and common English loanwords (backpacker) in their study on Dutch. Acronyms are usually much shorter than the full noun (GDP has only three syllables; *guomin shengchan zongzhi* has 6. EMBA and its Chinese translation have an even larger difference). In contrast, the English loanwords studied in Zenner et al. (2014) are just common words whose length is not very different from the Dutch word (*rugzakker* has 9 letter; *backpacker* has 10; both have 3 syllables). This could be the reason why Chinese speakers' preference of English acronyms is more closely related to the length of the word.

### Predictor 2: Freq

The linear regression model and Figure 3 clearly demonstrate an inverse proportional relation between concept frequency and the percentage of English. As the frequency of concept goes up, the use of English acronyms decreases. Therefore, it seems reasonable to say that the previous results of the three newspapers (briefly mentioned in Section 3.2.1) show a different trend simply because there are too few concepts in dataset 1. The generalization in Zenner et al (2014) still holds according to Figure 3. That is, if a concept is more frequent and thus more entrenched than others, it will be more likely to be used in the native language. Thus for GDP and WTO, only around 25% of the time they appear in English. That means most of the time readers of Xinhua News are seeing the Chinese words.

Therefore, as these concepts settle in into Chinese, their frequency is likely to increase each year. They will be more and more entrenched in the lexicon of Chinese speakers. In the end, all these concepts are likely to be used increasingly in Chinese. But we see NBA, DNA and MBA still have a large proportion of English usage, which suggests that frequency alone cannot explain everything. Predictors such as language economy also need to be counted.

*Figure 2* PercentOfEn ~ NumOfChar in Xinhua News corpus

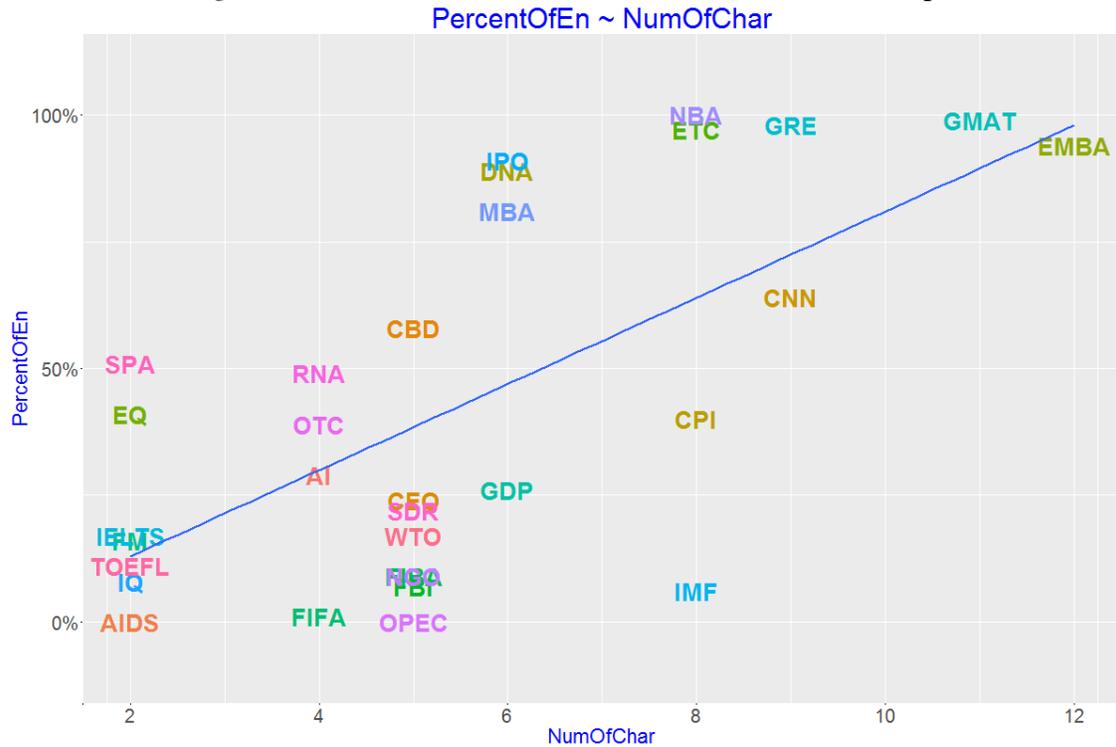

### Predictor 3: FirstAppear

Figure 4 shows that for all concepts in which English entered *PD* first (EN first, i.e. in the middle of Figure 4), the percentage of English acronyms is unanimously above 90%. On the other hand, for concepts whose Chinese entered *PD* first (CHN first, on the right of Figure 4), they have a lower percentage of English mostly. And in cases where English and Chinese entered *PD* at almost the same time (EN=CHN)[9], the data shows a random distribution.

Figure 4 suggests that the corpus of *PD* can somehow magically predict whether one foreign concept will be used more as an English acronym or Chinese. One reason might be that *PD* is the most authoritative newspaper in mainland China, and when in doubt, other newspapers are likely to consult *PD* and see how *PD* wrote about the concept—English or Chinese translation. This is suggested in the interview with the editor from *PD*, as she pointed out that reporters and editors tend to search for "precedence" of how certain concept is expressed and then follow suit. Since no news agency can be more authoritative than *PD*, it is natural for other newspapers to follow *PD*.

---

[9] This is operationalized in the following way: if the difference between the year of first English form and first Chinese form is fewer than 5 years, then they are considered to be entering the *PD* at the same time, i.e. EN=CHN

*Figure 3* PercentOfEn ~ Freq in Xinhua News corpus

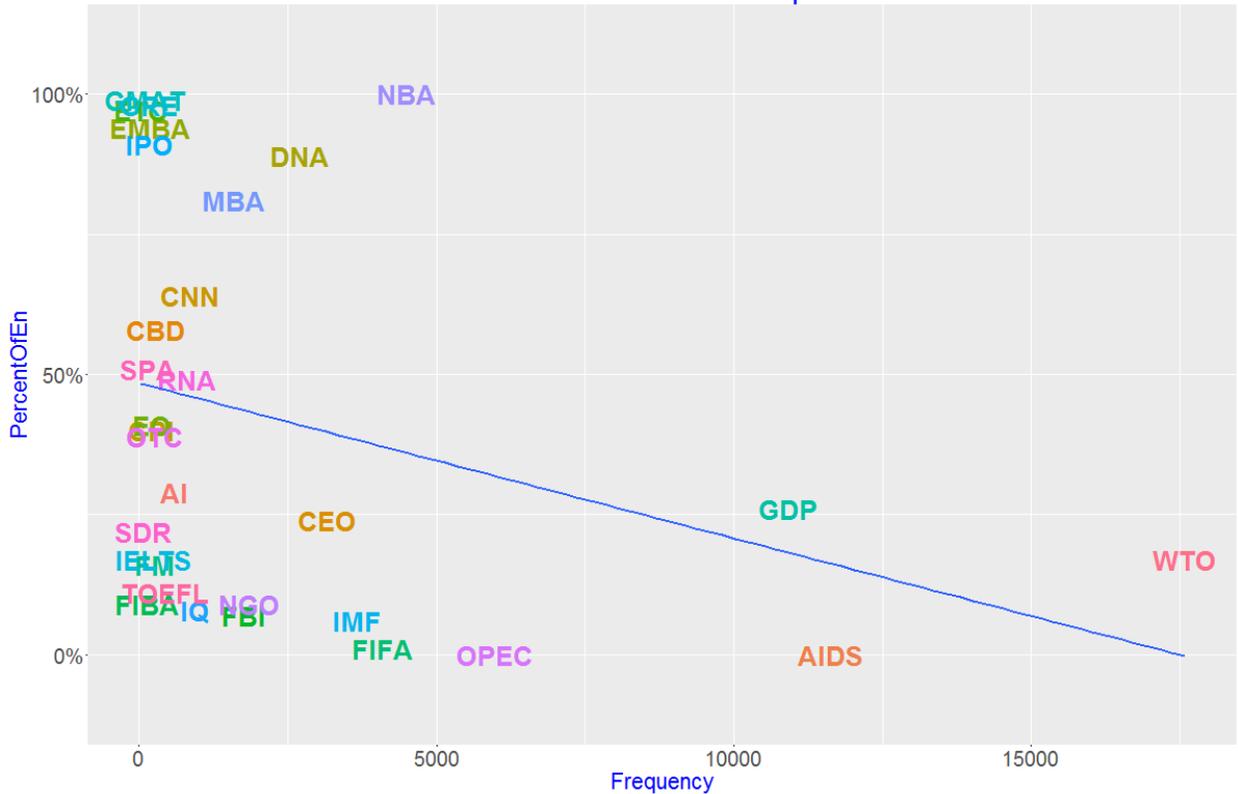

**Predictor 4: ProperN**

Inspired by the finding from dataset 1, I added a predictor that examines whether being a proper noun will influence *PercentOfEn*. Figure 5 has 8 panels, each showing concepts with 2, 4, 5, etc. number of characters in their translations. In each panel, acronyms on the left are not proper nouns, whereas those on the right are. It seems for acronyms with *NumOfChar* < 5, the assumption that proper nouns have lower *PercentOfEn* holds, but there are either no clear trend or not enough data for other cases. Crucially, in the linear regression model no significant effect is found for *ProperN*. Even if it is true that Chinese translations are preferred for proper nouns, I still don't have a convincing explanation at the moment.

**Predictor 5: ConceptAge**

The hypothesis for *ConceptAge* is that the older the concept, the more entrenched it should be in the native language lexicon, thus making it more likely for speakers to use the native (here Chinese) form (Zenner et al. 2014). However, in this study *ConceptAge* is not playing a significant role as shown in the results for the regression model in Table 5, which is different from the conclusion of Zenner and her group. One possibility is that in their Dutch study, the difference in concept age can be several hundred years, whereas in this study, the oldest concepts are only around 70 years old. That is, all acronyms are young in this study.

*Figure 4* PercentOfEn ~ FirstAppear in Xinhua News corpus

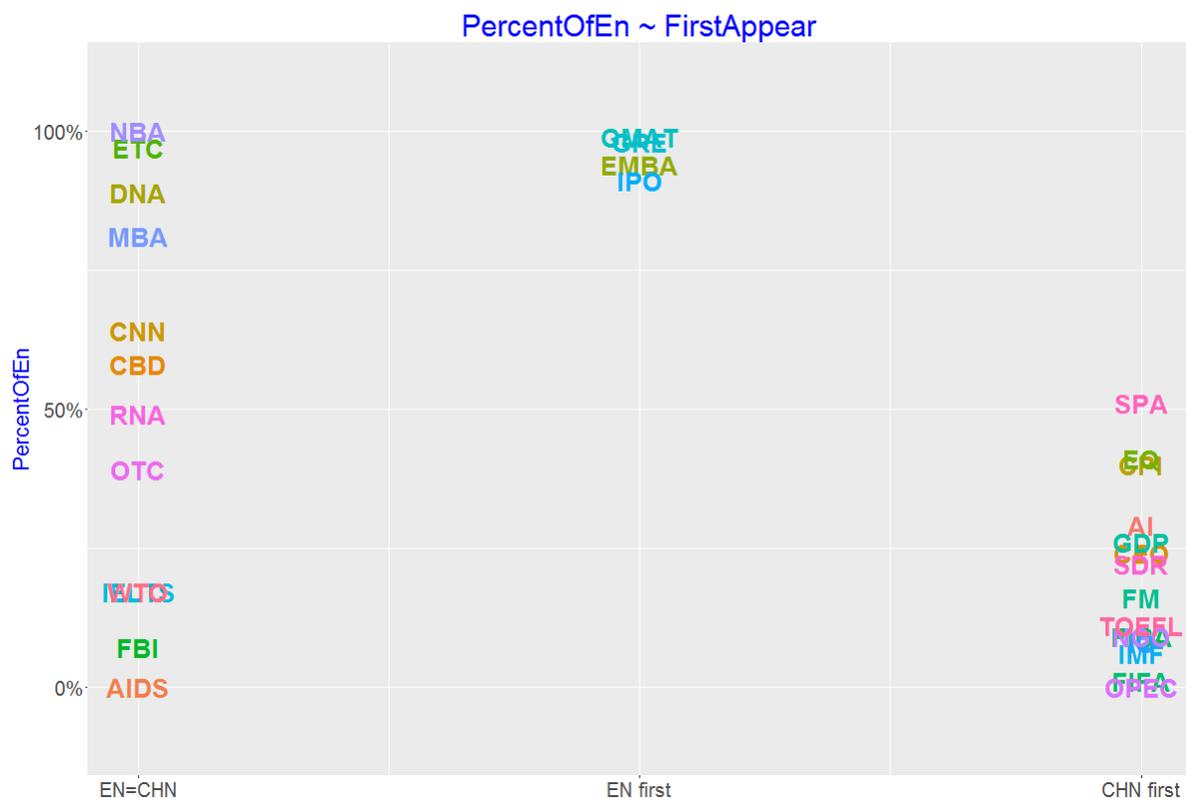

### 3.3 Interview with *PD* editor

In order to understand the role of language policy and also why *PD* has the lowest rate of English acronyms, the author had an unstructured interview with an editor from *PD*. According to her, *PD* has stricter rules than other newspapers about what to write when both an English acronym and its Chinese translation are available. For titles and headlines, only one English word—GDP—is allowed; all other English acronyms are strictly banned. This is confirmed in the data collection process, because for other newspapers we can always find articles with English words/acronyms in article titles, whereas for *PD*, English in the title is extremely rare.

The editor also suggested that for the content of the news articles, it is advised to use Chinese rather than the English acronym. One reason is the language policy we mentioned in Section 2.2.2. Another reason is that some loyal readers of *PD* have complained that they could not understand the English acronyms that appear in the newspaper (usually new and foreign concepts such as *Facebook, YouTube* or *Airbnb*). Therefore, English words and acronyms should be avoided or at least be accompanied by Chinese translations, which is the common practice. It is worth mentioning that the readership of *PD* is generally the middle-aged or older. Thus they may not be familiar with the new terms such as CEO or IPO. All the above reasons have contributed to the lower *PercentOfEn* in *PD*.

*Figure 5* PercentOfEn ~ ProperN in Xinhua News corpus

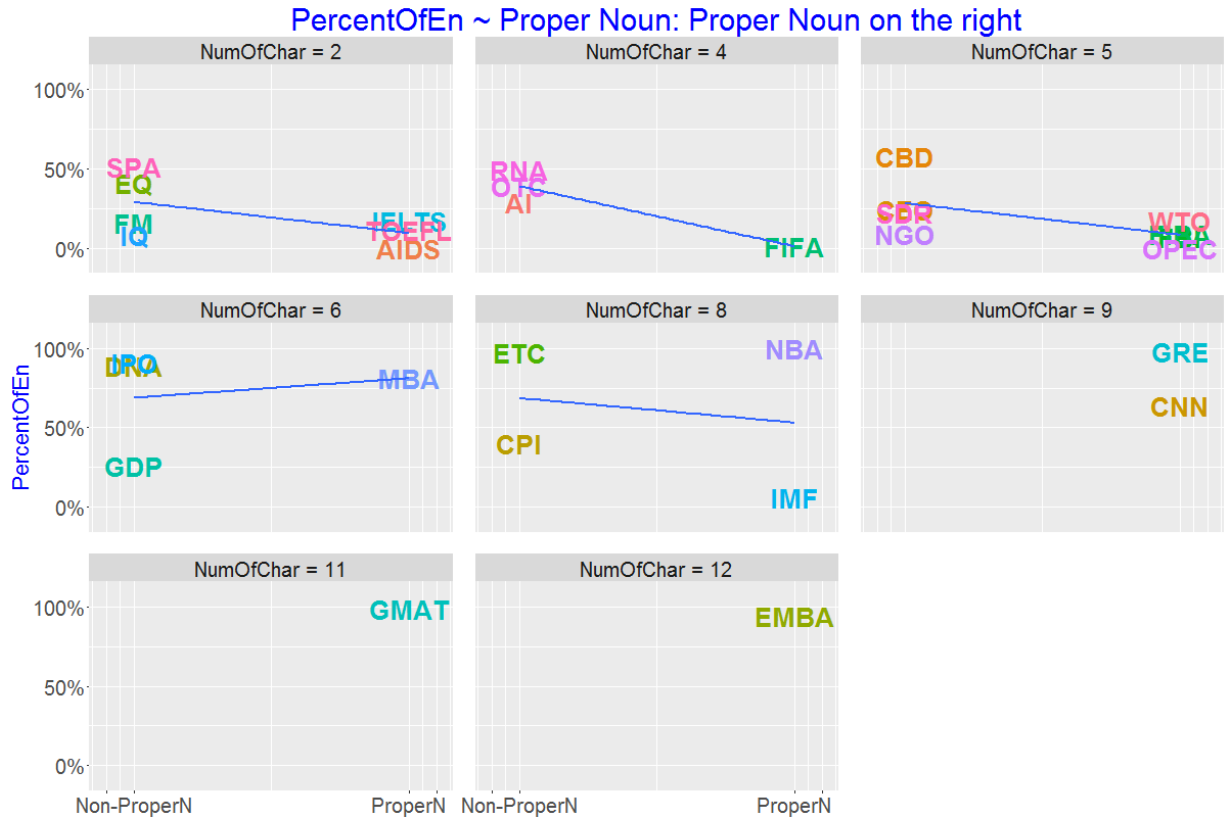

## 4. Conclusion

To sum up, this study has examined the use of English acronyms in Chinese newspapers by looking at the percentage of English *PercentOfEn* for different concepts. Results from two datasets show that different concepts have a varying *PercentOfEn*, ranging from 2% to 98%. Newspapers with more formal language (e.g. *People's Daily*) has a lower *PercentOfEn*.

Linear models in two datasets show that three predictors can predict *PercentOfEn* (all $p < .05$). That is, how much English is used for one concept can be predicted by 1) how often the concept occurs, 2) how long the Chinese translation is, and 3) whether the English or Chinese form first appeared in the newspaper. Language users choose the English form either because the Chinese translation is too long, or because the concept is not yet well entrenched in the Chinese lexicon, i.e. it may still need more time to "settle in".

This is one of the first quantitative studies that has systematically investigated the degree of English acronym usage in Chinese texts. It has implications for studies in language contact between Chinese and English and code-switching in Chinese written texts. Future studies can expand to more acronyms in other genres of written as well as spoken text, rendering a more comprehensive picture of the English acronym usage in Chinese.

APPENDIX
31 Concepts Used for Dataset 2[10]

| English acronym | English (full) | Chinese translation | Number of Characters |
|---|---|---|---|
| AIDS | Acquired Immune Deficiency Syndrome | 艾滋/ 获得性免疫缺陷综合征 | 2 |
| EQ | Emotional Quotient | 情商 | 2 |
| FM | Frequency Modulation | 调频 | 2 |
| IELTS | International English Language Testing System | 雅思 | 2 |
| IQ | Intelligent Quotient | 智商 | 2 |
| SPA | Solus Par Agula (Health by Water) | 水疗 | 2 |
| TOEFL | Test of English as a Foreign Language | 托福 | 2 |
| AI | Artificial Intelligence | 人工智能 | 4 |
| FIFA | Federation Internationale de Football Association | 国际足联/国际足球联盟 | 4 |
| OTC | Over-the-counter | 非处方药 | 4 |
| RNA | ribonucleic acid | 核糖核酸 | 4 |
| CBD | central business district | 中央商务区 | 5 |
| CEO | chief executive officer | 首席执行官 | 5 |
| FBI | Federal Bureau of Investigation | 联邦调查局 | 5 |
| FIBA | Fédération Internationale de Basketball Amateur | 国际篮联/国际篮球联合会 | 5 |
| NGO | non-governmental organization | 非政府组织 | 5 |
| OPEC | Organization of the Petroleum Exporting Countries | 欧佩克 / 石油输出国组织 | 5 |
| SDR | Special Drawing Rights | 特别提款权 | 5 |
| WTO | World Trade Organization | 世贸组织 / 世界贸易组织 | 5 |
| DNA | deoxyribonucleic acid | 脱氧核糖核酸 | 6 |
| GDP | gross domestic product | 国内生产总值 | 6 |
| IPO | initial public offering | 首次公开募股 | 6 |
| MBA | Master of Business Administration | 工商管理硕士 | 6 |
| CPI | consumer price index | 居民消费价格指数 | 8 |
| ETC | Electronic Toll Collection | 电子道路收费系统 | 8 |
| IMF | International Monetary Fund | 国际货币基金组织 | 8 |

[10] For an acronym that has multiple Chinese translations, I set numOfChar to the most frequent translation (as in AIDS) or the average of different translations (as in WTO, OPEC, FIBA). This may not be the best way to do it, but for this initial study I'll settle with this method. On hindsight, I realized that SPA, FIFA, FIBA are not English acronyms, which should have been excluded.

| NBA | National Basketball Association | 美国职业篮球联赛 | 8 |
|------|------|------|------|
| CNN | Cable News Network | 美国有线电视新闻网 | 9 |
| GRE | Graduate Record Examination | 美国研究生入学考试 | 9 |
| GMAT | Graduate Management Admission Test | 经企管理研究生入学考试 | 11 |
| EMBA | Executive Master of Business Administration | 高级管理人员工商管理硕士 | 12 |